%% file: main.tex
\documentclass[journal]{IEEEtran}
\usepackage{graphicx}
\usepackage[dvipsnames]{xcolor}

\ifCLASSINFOpdf
\else
\fi
\usepackage{array}
\begin{document}
%

\title{SARAL-Bot:\\  Autonomous Robot for Strawberry Plant Care}
%
%
%

\author{Arif~Ahmed ~~   
        Ritvik~Agarwal ~~
        Gaurav~Srikar ~~
        Nathaniel~Rose ~~ 
        Parikshit~Maini ~~ 
        \\
        University of Nevada Reno}

%
%

\markboth{
2024 ASABE Students Robotics Challenge (Advanced), Anaheim, CA}%
{Systems and Algorithms for Robot Autonomy Lab (SARAL), University of Nevada, Reno (UNR)}
%



\maketitle

\begin{abstract}
\input{0_abstract}
\end{abstract}


%
\IEEEpeerreviewmaketitle

\input{1_Est_Need_and_Benefit}  
\input{2_Approach} 
\input{3_Defn_of_Design_Objectives} 
\input{4_Parts_List}
\input{5_Hardware_Desc} 
\input{6_Software}

\input{7_Appropriateness_of_Tests}

\input{8_Achievement_of_Objectives}

\appendices

\section*{Acknowledgment}
We would like to express our deepest gratitude to all those who have supported and contributed to the development of our robot for the 2024 ASABE Student Robotics Competition. First and foremost, we extend our sincere thanks to our faculty advisor, Dr. Parikshit Maini, for his unwavering guidance, expertise, and mentorship throughout the project. Their valuable insights and encouragement have been instrumental in helping us navigate the challenges and achieve our goals. We are grateful to the ASABE (American Society of Agricultural and Biological Engineers) for organizing this competition and providing a platform for students to showcase their skills and innovations in agricultural robotics. The competition has been a source of inspiration and motivation for our team, driving us to push the boundaries of our knowledge, skills and capabilities. We also extend our appreciation to the open-source community, particularly the developers and contributors of ROS2, RTABMAP, Moveit2, and other libraries and tools that have been integral to our robot's software stack. Their dedication to creating accessible and powerful software has greatly facilitated our development process.

\ifCLASSOPTIONcaptionsoff
  \newpage
\fi

\bibliographystyle{IEEEtran}
\bibliography{bibtex/bib/asabe}

\end{document}

%% file: 0_abstract.tex
Strawberry farming demands intensive labor for monitoring and maintaining plant health. To address this, Team SARAL develops an autonomous robot for the 2024 ASABE Student Robotics Challenge, capable of navigation, unhealthy leaf detection, and removal. The system addresses labor shortages, reduces costs, and supports sustainable farming through vision-based plant assessment. This work demonstrates the potential of robotics to modernize strawberry cultivation and enable scalable, intelligent agricultural solutions.

%% file: 1_Est_Need_and_Benefit.tex
\section{Establishment of Need and Benefit to Agriculture} %
The 2024 ASABE Student Robotics Challenge aims to simulate the task of strawberry farming, which presents an excellent opportunity to demonstrate the potential benefits of robotics in agriculture. Strawberries are extensively consumed worldwide and the Food and Agriculture Organization (FAO) reported the increasing production of strawberries worldwide \cite{park2021design}.  Strawberry production is labor-intensive, requiring significant manual effort for tasks such as planting, monitoring plant health, removing unhealthy leaves, and harvesting. These tasks can be time-consuming, costly, and physically demanding for human workers.
The development of autonomous robots capable of navigating strawberry fields, identifying healthy and unhealthy plant parts, and precisely removing unhealthy leaves and flowers has the potential to revolutionize the strawberry farming industry. By automating these tasks, farmers can:
\begin{enumerate}
    \item Reduce labor costs: Autonomous robots can work continuously without the need for breaks, thereby reducing the number of human workers required and associated labor costs.
    \item Improve efficiency: Robots can complete tasks faster and more consistently than human workers, leading to increased overall efficiency in strawberry production.
    \item Enhance crop health: Robots equipped with advanced sensors can accurately identify unhealthy plant parts and remove them promptly, preventing the spread of diseases and ultimately improving crop health and yield.
    \item Enable precision agriculture: Robotics allows for targeted treatments and interventions, reducing the use of resources such as water, fertilizers, and pesticides, thus promoting sustainable agricultural practices.
    \item Alleviate labor shortages: As the agricultural industry faces increasing labor shortages, robots can help fill the gap and ensure that critical tasks are completed on time.
\end{enumerate}

By participating in the 2024 ASABE Student Robotics Challenge, our team aims to contribute to the development of innovative robotic solutions that address the needs of the strawberry farming industry. Our robot demonstrates the potential for autonomous systems to improve efficiency, crop health, and sustainability in agriculture, ultimately benefiting farmers, consumers, and the environment.

In order to satisfy the strawberry market demands, it is obvious that we need to increase the production of strawberries. Recent robotics research on agriculture reported that robotic solutions can help monitor plants and increase the overall yield of strawberries \cite{defterli2016review, ren2023mobile}. To get a high yield of strawberries it is important to identify if the plants are suffering from nutrient deficiencies, diseases, and/or incorrect soil pH. Interestingly, only by inspecting the leaf-color it is possible to identify several factors.\footnote{https://www.epicgardening.com/strawberry-problems/}  For example, the yellow or pale leaves indicate that the plant had moisture stress due to over-watering. When the leaves start to show such characteristics, it is necessary to prune those unhealthy leaves so that the symptoms do not spread to other leaves. Similarly, there are some symptoms such as fungal attack, airflow problem, cold snap issue etc. Those problems can be easily identifiable using the robotic solutions \cite{ren2023mobile}.

\begin{figure}[!htp]
    \centering
    \includegraphics[width=\linewidth]{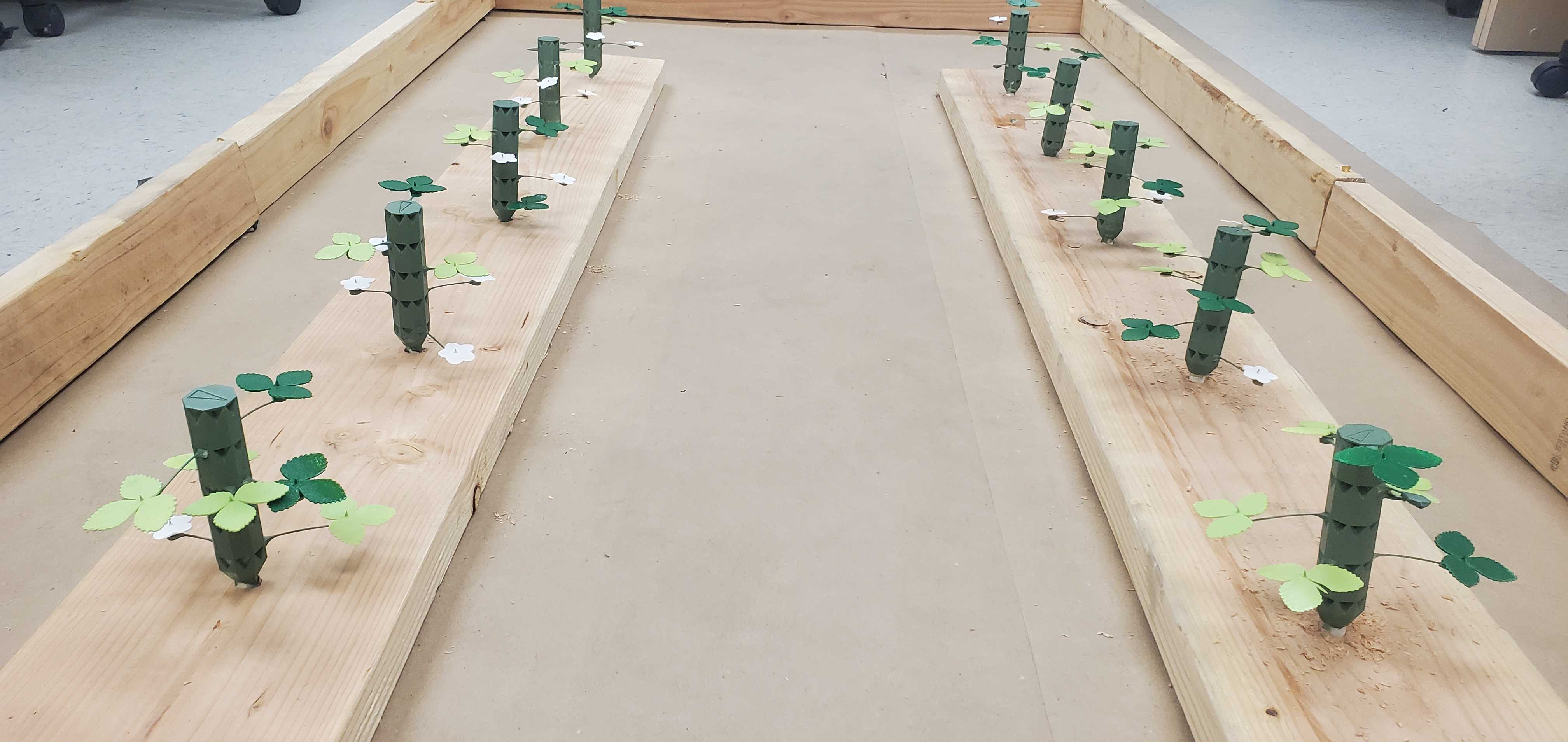}
    \caption{Simulated ASABE 2024 Competition Arena (preparatory stage) }
    \label{fig:arena}
\end{figure}

%% file: 2_Approach.tex
\section{Approach and Originality}  
\label{sec:approach}
Our team has developed an innovative solution for the advanced division of the 2024 ASABE Student Robotics Challenge, leveraging the capabilities of an off-the-shelf robot, the HiWonder ArmPi Pro \footnote{https://www.hiwonder.com/products/armpi-pro}, and building on top of it with custom hardware and software.

\subsubsection{Hardware Enhancements}
To improve the robot's localization capabilities while adhering to the size constraints, we have mounted an Intel D435i \footnote{https://www.intelrealsense.com/depth-camera-d435i/} RGBD camera on a linear actuator at the rear of the ArmPi Pro, facing forward. 
The linear actuator allows the depth camera (i.e., D435i RGBD camera) have better and increased field of view. 
Moreover, this design configuration also helps us perform a robust Visual Inertial Odometry (VIO) 
using a VIO algorithm called Real‐Time Appearance‐Based Mapping (RTABMAP) \cite{labbe2019rtab}.

\subsubsection{Software Architecture}
Our software stack is built on ROS2 Humble, the latest LTS version, ensuring compatibility and stability.\footnote{https://docs.ros.org/en/humble/index.html} We have developed a modular architecture consisting of three main systems (all in ROS2 Humble): 
\begin{enumerate}
    \item Behavior Coordinator - responsible for task sequence
    \item Navigation System - responsible for mobility control
    \item Plant Processing System - responsible for plant trimming
\end{enumerate}
These subsystems then communicate with the provided ArmPi Pro's control nodes via a ROS1 Node made by us which exposes various REST API endpoints via a webserver. Finally, in order to ensure reliable and repeatable deployment, we use Docker for environment management\footnote{https://www.docker.com/}. 

\subsubsection{Robot Behaviour and Task Strategy}
The robot behaviour is focused on coordinating our navigation and plant processing systems. Given the static nature of the arena, we have optimized our navigation strategy by limiting planning to pre-defined obstacle-free straight-line movements.

\begin{figure}
  \begin{center}
  \includegraphics[width=0.99\linewidth]{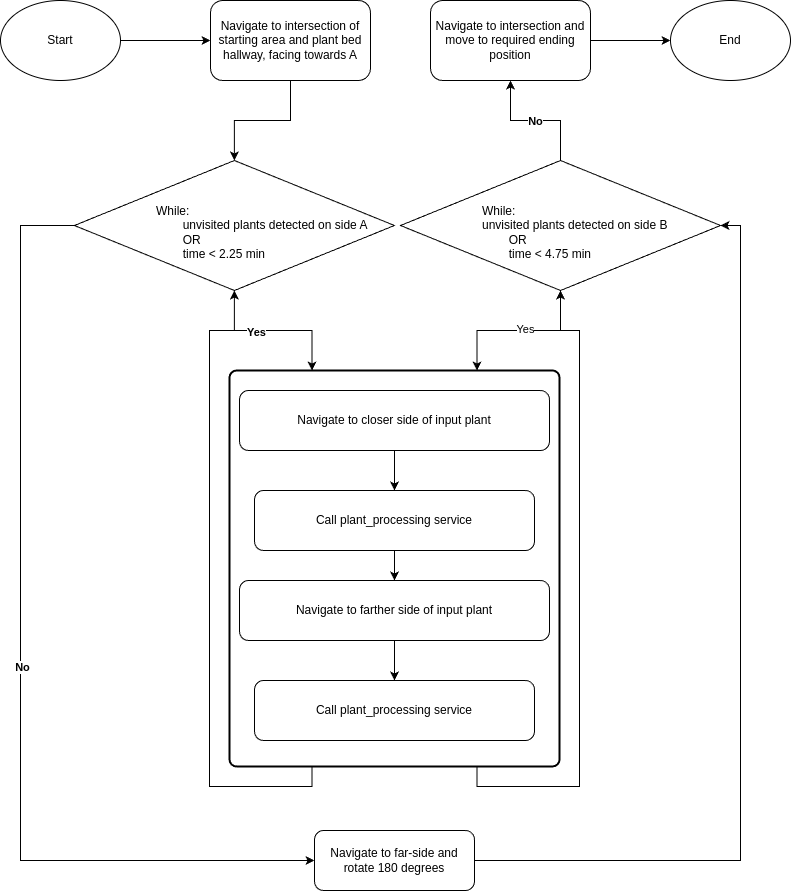}
  \caption{Flowchart of SARAL-Bot Behaviour Coordinator Logic Coordinator}
  \label{fig:beh_flow}
\end{center}
\end{figure}

\subsubsection{Arm Control and Inverse Kinematics}
For controlling the ArmPi Pro's chassis and arm, we utilize the onboard Raspberry Pi, running a Flask server with a ROS Melodic node. This allows seamless integration of the existing hardware with our ROS2-based software stack.
Inverse kinematics for the robotic arm is handled using MoveIt 2, a powerful motion planning framework. This enables our robot to accurately reach leaves and flowers at varying heights and distances.

\subsubsection{Computer Vision Techniques}
To detect and differentiate between healthy leaves, unhealthy leaves, and flowers, we employ HSV-based color thresholding followed by contour detection. This computer vision technique ensures robust performance under varying lighting conditions.



Our unique combination of hardware enhancements, modular software architecture, optimized navigation strategy, and advanced computer vision techniques demonstrates our team's innovative approach to tackling the challenges presented by the competition.

%% file: 3_Defn_of_Design_Objectives.tex
\begin{figure}[!htp]
    \centering
    \includegraphics[width=\linewidth]{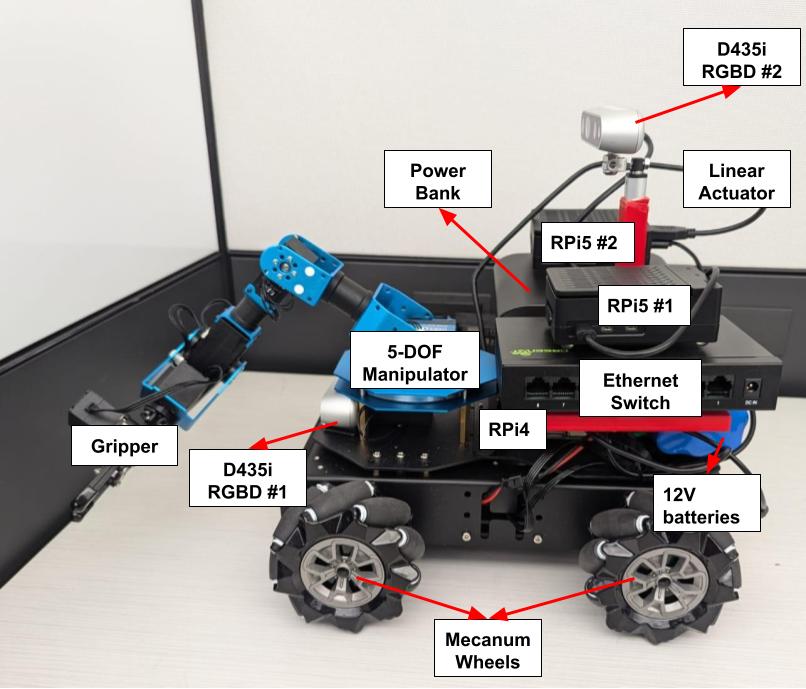}
    \caption{
    The Robot Design: Stacking various hardware components to handle specific computational tasks. In this image, an RPi4 is mounted directly on top of the robot chassis, with a power bank, Ethernet switch, and two RPi5 units stacked on top of it.}
    \label{fig:robot_design_1}
\end{figure}
\section{Definition of Design Objectives and Criteria } 
\label{sec:objectives}
Our team established clear design objectives and criteria to guide the development of our robot for the 2024 ASABE Student Robotics Challenge. These objectives were based on a thorough understanding of the competition rules, constraints, and challenges, ensuring that our robot would be well-suited to perform the required tasks effectively.
\subsubsection{Alignment with Competition Requirements}
Our primary design objective was to develop a robot capable of autonomously navigating the arena, inspecting plants, and performing selective harvesting while adhering to the size constraints specified in the competition rules. We aimed to create a compact and efficient design that could complete the tasks within the allotted time frame.
\subsubsection{Performance Goals}
We set specific performance goals for our robot, including:
\begin{itemize}
\item Achieving a navigation speed of at least 10 cm per second
\item Maintaining a localization accuracy within 2 centimeters of the ground truth
\item Detecting plant health with an accuracy of 95 percent or higher
\item Successfully harvesting at least 90 percent of the target leaves and flowers
\end{itemize}
These quantitative goals helped us focus our efforts and evaluate the success of our design.
\subsubsection{System Architecture and Component Selection}
Our design objectives influenced our choice of system architecture and hardware components. We selected the ArmPi Pro as our base robot for its holonomic navigation capabilities, and we added an Intel D435i RGBD camera on a linear actuator to enhance localization and object detection. We chose ROS2 Humble as our software framework for its stability, modularity, and community support.

\subsubsection{Modularity, Scalability, and Ease of Use}
We prioritized modularity and scalability in our design, allowing for easy integration of new features and algorithms as the project progressed. We also aimed to create a user-friendly system that could be easily set up, operated, and maintained by our team members. 

\subsubsection{Reliability and Robustness}
Ensuring the reliability and robustness of our robot was a critical design objective. We aimed to develop a system that could perform consistently and handle unexpected situations gracefully. This involved extensive testing and validation of our hardware and software components. For example, to increase the computational power of our system while adhering to the imposed budget, we decided to go for a distributed network of Raspberry Pis rather than a single higher-end computer.

\subsubsection{Cost and Resource Optimization}
Finally, we sought to optimize cost and resource utilization throughout the design process. We carefully selected components that balanced performance and affordability, and we leveraged open-source software and libraries to reduce development time and cost.
By defining clear and specific design objectives and criteria, we established a strong foundation for the development of our robot, ensuring that it would be well-equipped to tackle the challenges of the 2024 ASABE Student Robotics Challenge.

%% file: 4_Parts_List.tex
\section{Parts List and Table} 
We show the parts list and prices in Table~\ref{tab:bom} in the bill of materials. The table categorizes components (robot, computational resources, sensors, power supplies, 3D printed parts and miscellaneous items) and gives the details for each component within a category. We've also emphasized that the total compute resources used on the robot adhere to the imposed control components budget, i.e. \$ 500.00.







\begin{table}[!htbp]
\centering
\begin{tabular}{l|r}
\hline
Item & Cost Per Unit \\
\hline
\multicolumn{1}{c}{\textbf{HiWonder Arm Pi Pro}} &  \textbf{\$ 529.00} \\
1 $\times$ Aluminum alloy chassis (256mm $\times$ 298mm) &  \\
4 $\times$ Aluminum mecanum wheels (50mm $\times$ 100mm) &  \\

1 $\times$ LX-225 servo (included in ArmPi Pro) & \\
3 $\times$ LX-15D servo  (included in ArmPi Pro)  & \ \\
1 $\times$ ID 1 servo  (included in ArmPi Pro) & \\
1 $\times$ Gripper  (included in ArmPi Pro)  &  \\
4 $\times$ Motor encoders &   \\
1 $\times$ Arm Pi Pro 7.4V lithium ion battery (6000mAh) &  \\
1 $\times$ Raspberry Pi 4 (included in ArmPi Pro)  &    \\
\hline
\multicolumn{2}{c}{\textbf{Added Computational Resources}} \\
2 $\times$ Raspberry Pi 5 &  \$ 90.00 \\
1 $\times$ TP-Link 5 port Ethernet switch & \$ 18.00 \\
\hline
\multicolumn{2}{c}{\textbf{Sensors}} \\
2 $\times$ Intel D435i depth camera & \$ 327.00 \\
\hline
\multicolumn{2}{c}{\textbf{Power Suppliers}} \\
1 $\times$ KBT 12V lithium ion battery (2400mAh) & \$ 23.00 \\
1 $\times$ Krisdonia 5V power bank (25000mAh) & \$ 89.50 \\
\hline
\multicolumn{2}{c}{\textbf{Added Hardware}} \\
1 $\times$ 12V Linear actuator(180mm length w/125mm stroke) & \$ 28.00 \\
1 $\times$ 12V 2-Channel Relay Switch & \$ 8.99 \\

\multicolumn{2}{c}{\textbf{Custom 3D Printed Parts}} \\
1 $\times$ 3D printed chassis back (225.5mm$\times$112mm$\times$140mm) & \$ 18.80 \\
1 $\times$ 3D printed custom gripper (100mm length) & \$ 0.94 \\
\hline
\multicolumn{2}{c}{\textbf{}} \\
\multicolumn{1}{c}{\textbf{TOTAL COMPUTE RESOURCES}} &  \textbf{\$ 260.00} \\
\multicolumn{2}{c}{\textbf{}} \\
\end{tabular}
\caption{Bill of Materials with Prices}
\label{tab:bom}
\end{table}

\begin{figure}[!htp]
    \centering
    \includegraphics[width=0.76\linewidth]{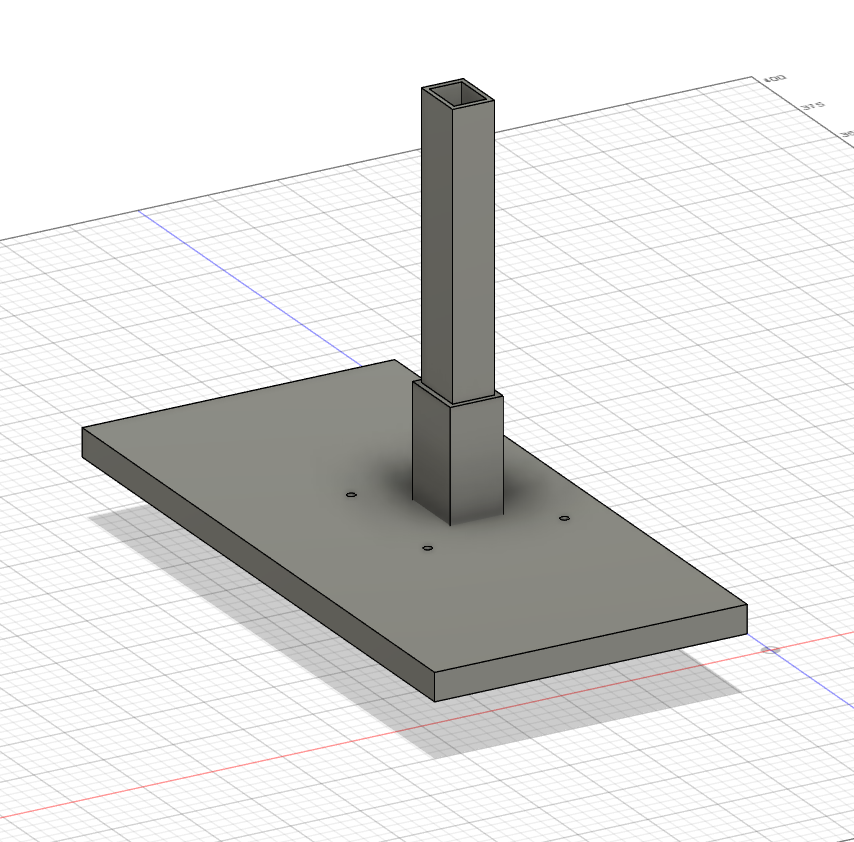}
    \caption{Custom Chassis Back: This custom design is attached to the back of our aluminum chassis in order to increase the space we can utilize for components and to securely hold the linear actuator in place. The custom back is 10mm thick, 225.50mm wide, and 112.00mm long.}
    \label{fig:3D_Print_Back}
\end{figure}

\begin{figure}[!htp]
    \centering
    \includegraphics[width=0.76\linewidth]{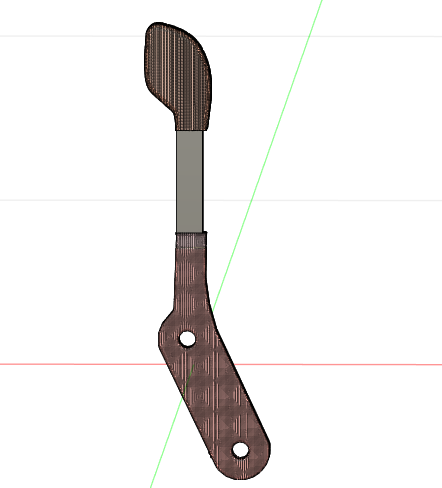}
    \caption{Custom Gripper: This custom gripper design is utilized to extend the reach of our original gripper from 60mm to 100mm.}
    \label{fig:3D_Print_Gripper}
\end{figure}

%% file: 5_Hardware_Desc.tex
\section{Hardware Description} 

\subsection{Chassis and the Arm: HiWonder Arm Pi Pro}
The ArmpiPro chassis is fully manufactured with a hard aluminum alloy to keep it durable while also being lightweight. The chassis is 256mm wide, 298mm long, and 148mm tall, which puts it within the boundaries of the competition's size limitations. The chassis has several threaded holes throughout to allow for easy customization and easy wire management. The SARAL Bot's chassis has been customized with the use of a custom 3D printed back to stabilize the linear actuator while also adding space for additional components. New Mecanum wheels have been chosen to replace the initial ones that come with the Arm Pi Pro. These new wheels are made of aluminum alloy, minimizing risks of damage, are heavier, meaning they provide better stability to the mobile platform, and have fully rubber rollers, allowing for better traction and therefore control while traversing the peat moss filled terrain.

\begin{figure}[!htp]
    \centering
    \includegraphics[width=0.76\linewidth]{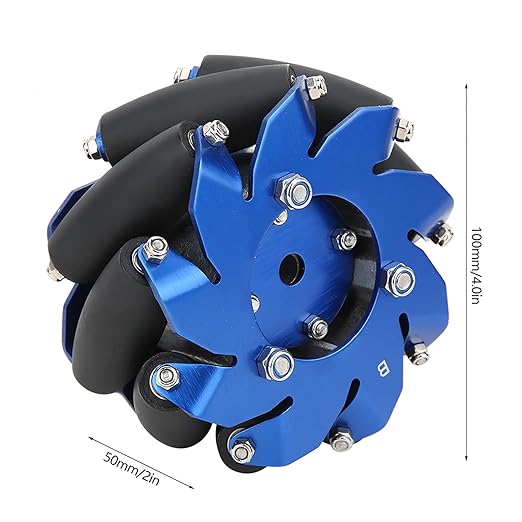}
    \caption{Aluminum Mecanum Wheels: These new mecanum wheels are fully manufactured from aluminum alloy, making them more durable and heavier than the original plastic mecanum wheels. They also utilize fully rubber rollers allowing for better traction. }
    \label{fig:Mechanum Wheels}
\end{figure}

The manipulator mounted on our saral\_bot features a 5 Degree of Freedom arm, powered by servo motors. Initially, the orientations of each joint in the arm were configured to perform pitch operations, except for the 1st and 5th joints, which were yaw-based, and the end effector. This configuration limited the arm's reach, particularly in accessing unhealthy leaves located at the back of a plant model. To enhance its reach, we altered the orientation of the 4th joint to perform rolling operations. This adjustment enables the arm to extend around either side of a plant, bend the forward arm sideways, and precisely reach the leaves with the gripper. The modified orientation configuration of the arm is detailed in table \ref{tab:servo_specs}.

The Armpi Pro robot's built-in gripper was too short to reach leaves behind the plant without disturbing the surrounding healthy leaves. We developed a longer fork gripper to grasp unhealthy leaves effectively without impacting the adjacent foliage. The designs of these grippers are depicted in Fig. \ref{fig:3D_Print_Gripper}. The initial reach of the arm was 363mm. With the extended gripper, the new total reach is 403mm.

The Armpi Pro robot was originally equipped with a built-in camera located on the 5th link of the arm. However, for this challenge, an eye-in-hand configuration was not ideal and was effectively reducing the power output of the end-effector, so we removed the pre-mounted camera.

When not in use for manipulation, the manipulator is positioned in a default pose to avoid obstructing the sensing perspective from the top of the linear actuator.

\subsection{Computational Resources and Networking}
Our robot's computational resources consist of three Raspberry Pi (RPi) single-board computers: two RPi 5 models and one RPi 4 which camer integrated with the ArmPi Pro. This distributed computing architecture allows us to efficiently handle the various computational tasks required for the robot's operation.
The two RPi 5 models are dedicated to running our custom ROS2 nodes, which handle high-level decision making, navigation, and plant processing. These nodes communicate with each other using ROS2's built-in communication infrastructure, leveraging the publish-subscribe and service-client patterns. The RPi 5 models were chosen for their enhanced processing power and improved I/O capabilities compared to previous Raspberry Pi generations, ensuring smooth and responsive performance of our ROS2 system while adhering to the compute budget.
The RPi 4 model, which comes integrated with the ArmPi Pro, is responsible for running the low-level control software for the robot's motors and servos. This control software, based on ROS1 Melodic, communicates with the higher-level ROS2 nodes running on the RPi 5 models through a custom ROS1-ROS2 bridge implemented using a Flask web server. This is further discussed in the ROS1-Flask Server subsection in Section \ref{sec:sw_ov}. 
To facilitate communication between the three RPi boards, we employ a TP-Link 5-port Ethernet switch. The switch establishes a local area network (LAN) connecting the RPis via IPs configured by us, enabling them to exchange data and messages efficiently. The choice of a wired Ethernet connection, as opposed to wireless communication, ensures reliable and low-latency communication between the computational resources, which is critical for real-time robot control and coordination.
By distributing the computational workload across multiple RPi boards and establishing a robust wired network for communication, we have created a powerful and flexible computational infrastructure for our robot. This architecture allows us to efficiently process sensor data, perform complex calculations, and coordinate the robot's various subsystems, ultimately enabling it to navigate, inspect plants, and perform selective harvesting tasks with high accuracy and reliability.

\subsection{Sensors}

We are using two Intel D435i depth cameras for visual perception (See Fig. \ref{fig:robot_design_1}). The D435i is a high-performance, low-power stereo vision camera that provides accurate depth information. It features a pair of depth sensors, an RGB sensor, and an infrared projector, enabling it to capture high-resolution 3D images in various lighting conditions. One camera is mounted on a linear actuator at the rear end of the robot which will expand at the start. This placement provides a more isometric view of the plants while allowing the arm to function without occluding them. The other RGDB camera is mounted in the front and enables the robot to perform accurate localization and mapping.

In addition, the motor encoders on wheels and the arm servos are being used to achieve the desired outputs.  Three different types of servos are utilized: LX-225, LX-15D, and ID 1 servo. More details on the servo motors are described in table \ref{tab:servo_specs}, as well. All these servos perform the manipulation with $0.3^{\circ}$ accuracy.

\begin{table}[h!]
\centering
\begin{tabular}{|>{\centering\arraybackslash}m{0.8cm}|>{\centering\arraybackslash}m{1cm}|>{\centering\arraybackslash}m{1.2cm}|>{\centering\arraybackslash}m{2.5cm}|>{\centering\arraybackslash}m{1cm}|}
\hline
\textbf{Joint ID} & \textbf{Servo Type} & \hspace{-0.1 cm} \textbf{Orientation} & \textbf{Torque} & \textbf{Limits {\tiny(In Radians)}} \\ \hline
Joint $1$ & LX-$15$D & Yaw & $15 kg/cm$ with $6V$ \newline $17 kg/cm$ with $7.4V$ & $-\frac{\pi}{2} to \frac{\pi}{2}$ \\ \hline
Joint $2$ & LX-$225$ & Pitch & $25 kg/cm$ with $7.4V$ & $-\frac{\pi}{2} to \frac{\pi}{2}$ \\ \hline
Joint $3$ & LX-$15$D & Pitch & $15 kg/cm$ with $6V$ \newline $17 kg/cm$ with $7.4V$ & $-\frac{\pi}{2} to \frac{\pi}{2}$ \\ \hline
Joint $4$ & LX-$15$D & Roll & $15 kg/cm$ with $6V$ \newline $17 kg/cm$ with $7.4V$ & $-\frac{\pi}{2} to \frac{\pi}{2}$ \\ \hline
Joint $5$ & ID $1$ & Yaw & $8 kg/cm$ with $7.4V$ & $-\frac{\pi}{2} to \frac{\pi}{2}$ \\ \hline
\end{tabular}
\vspace{0.3 cm} 
\caption{Servo Specifications for Robot Joints in "Zero State"}
\label{tab:servo_specs}
\vspace{-1.2 cm}
\end{table}

\subsection{Power Supply}
The system design utilizes 3 different power supplies due to the varying power requirements of different components. The first power supplier is provided with the Arm Pi Pro and is a 7.4V lithium ion polymer (Lipo) battery with 6000mAh which is used to power the 7.4V motors in the mobile platform and manipulator. The 7.4V Lipo battery also powers the RPi 4 through the use of a step down chip to bring it to the intended 5V voltage. The next power supplier is another Lipo battery that is 12V with 2400mAh and is used to power the linear actuator through the use of a 5V to 12V 2-channel relay switch which gives the correct 12V inputs to the linear actuator depending on the 5V inputs it receives, thus allowing precise control over the linear actuator, which is crucial to positioning the rear-mounted depth camera for plant inspection and manipulation. The final supplier is a 5V power bank with 25000mAh which powers both RPi 5 modules, both D435i depth cameras, and the Ethernet switch. 

%% file: 6_Software.tex
\section{Software Overview} 
\label{sec:sw_ov}
We select the Robot Operating System 2 (ROS2) Humble instead of ROS1 for all our sensing and planning tasks. Overall, ROS2 is a more reliable and high-quality framework compared to ROS1 \cite{ros2_design}. ROS2 uses highly secure DDS communication compared to the TCP/UDP used in ROS1. Besides, ROS2 supports multiple nodes per process, whereas ROS1 allows only a single node per process. However, the existing control software for the wheel and arm motors, provided by ArmPi Pro is based on ROS1 Melodic. To avoid reinventing the wheel, we built competition-specific software on top of this foundation. Our software architecture consists of three main modules: Behaviour Coordinator, Navigation System, and the Plant Processing System. These modules together achieve the desired robot functionality in the competition.

\subsection{ROS1-Flask Server}
In order to interface the ROS1 system and the ROS2 system built on top of it, we created a ROS1 Node that also runs an internal web server. This server exposes REST API endpoints which when hit by our ROS2 system with desired motor inputs as parameters, publishes the equivalent ROS1 messages to control the wheel and arm motors. 

We opted for the Flask REST API instead of a ROS2 bridge due to the lack of support for ROS2 (Dashing) on Ubuntu 18.04 running on the Armpi Pro’s Pi. Thus, the Flask REST API provides a reliable channel for communication between ROS1 and ROS2.

\subsection{Behaviour Coordinator}
The Behaviour Coordinator (as shown in Fig~\ref{fig:beh_flow}) is responsible for coordinating the overall robot behavior and managing the high-level decision-making process. It communicates with the Navigation System and Plant Processing System using custom ROS2 messages to trigger the appropriate actions based on the current state and sensor data. The Behaviour Coordinator implements the following high-level logic:
\begin{enumerate}
\item Navigate to the intersection of the starting area and plant bed hallway, facing towards plant bed A.
\item While there are unvisited plants detected on side A or the elapsed time is less than $\sim2.25$ minutes, navigate to the nearest side of the next detected plant on side A and invoke the Plant Processing System.
\item Navigate to the far side of the arena and rotate 180 degrees.
\item While there are unvisited plants detected on side B or the elapsed time is less than $\sim4.75$ minutes, navigate to the nearest side of the next detected plant on side B and call the Plant Processing System.
\item Navigate to the intersection and move to the required ending position.
\end{enumerate}
The front mounted camera on the robot provides the necessary data for robot localization with respect to the global frame. The rear-end front facing camera, which is elevated along the  vertical axis when the linear actuator is extended, is accessed by the Plant Processing System for visually analyzing the plants. The identified targets i.e. unhealthy leaves and unwanted flowers, and their 3D locations are then utilized to perform the arm positioning and the plucking maneuver.

\subsection{Navigation System}
\begin{figure}[!htp]
\centering
\includegraphics[width=\linewidth]{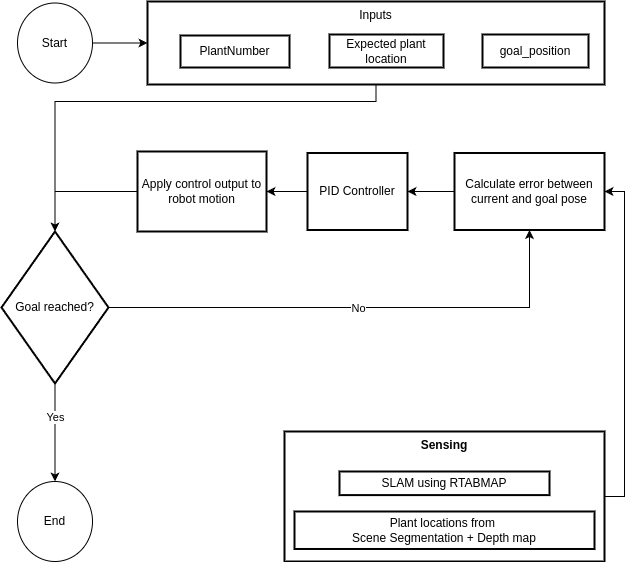}
\caption{Flowchart of SARAL-Bot Navigation System}
\label{fig:navigation_system}
\vspace{-0.5 cm}
\end{figure}

The Navigation System (see Fig. \ref{fig:navigation_system} ) is responsible for localizing the robot within the arena and planning paths to reach the desired goal positions. It relies on sensor data from the Intel RealSense D435i depth cameras and uses RTABMAP for visual SLAM (Simultaneous Localization and Mapping), thus constructing an accurate map of the environment. As the robot behaviour progresses, navigation goal pose is decided and conveyed by the Behaviour Coordinator, which ensures that there are no obstacles in the straight line path connecting the current pose to the goal pose. Now, a simple PID controller is used to minimize the error between the two poses. On completion the control goes back to the Behaviour Coordinator.

\subsection{Plant Processing System}
The Plant Processing System is activated by the Behaviour Coordinator when the robot is positioned near a plant. It utilizes the rear-end front-facing depth camera on the linear actuator to capture RGBD images of the plant. A vision pipeline to detect and localize the leaves and flowers of each plant is put in place. The pipeline transforms RGB into the HSV (Hue, Saturation, Value) color space and identifies healthy leaves, unhealthy leaves, and flowers based on the color thresholds derived from the information provided in the competition rules. To reduce false positives, a planarity check is performed using the RANSAC algorithm \cite{fischler1981random}, discarding any detected objects that are not planar.

\begin{figure}[!htbp]
\centering
\includegraphics[width=\linewidth]{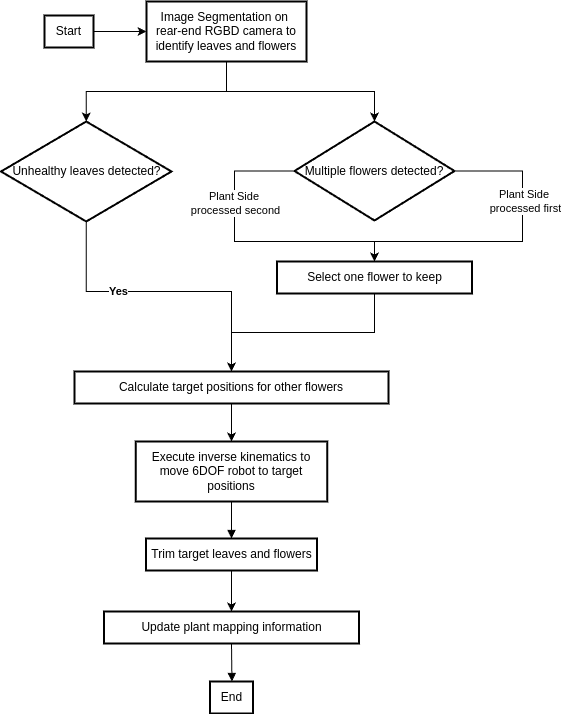}
\caption{Flowchart of the SARAL-Bot Plant Processing System }
\label{fig:plant_processing_system}
\end{figure}

The depth information from the camera is then used to localize the detected leaves and flowers with respect to the camera frame. After detecting, identifying, and localizing the plant parts, the `Plant Processing System' performs the following tasks:
\begin{enumerate}
\item Publish the counts and poses of the plant parts using a custom ROS2 message.
\item Transform the poses of the unhealthy leaves and flowers to the manipulator's base link.
\item Based on the input from the behaviour node, if we are processing the nearer side of the plant, we skip one flower from trimming. On the other hand, if we are processing the farther side of the plant, depending on whether there is 0 or 1 flower left on the nearer side, we trim all but one or all flowers, respectively.
\item Send manipulation commands to the ArmPi Pro's joint state control node using REST API requests, thus reaching the required pose for the modified gripper to trim the target.
\end{enumerate}

\begin{figure*}[!htb]
\centering
\begin{minipage}[b]{0.57\linewidth}
\includegraphics[width=\linewidth]{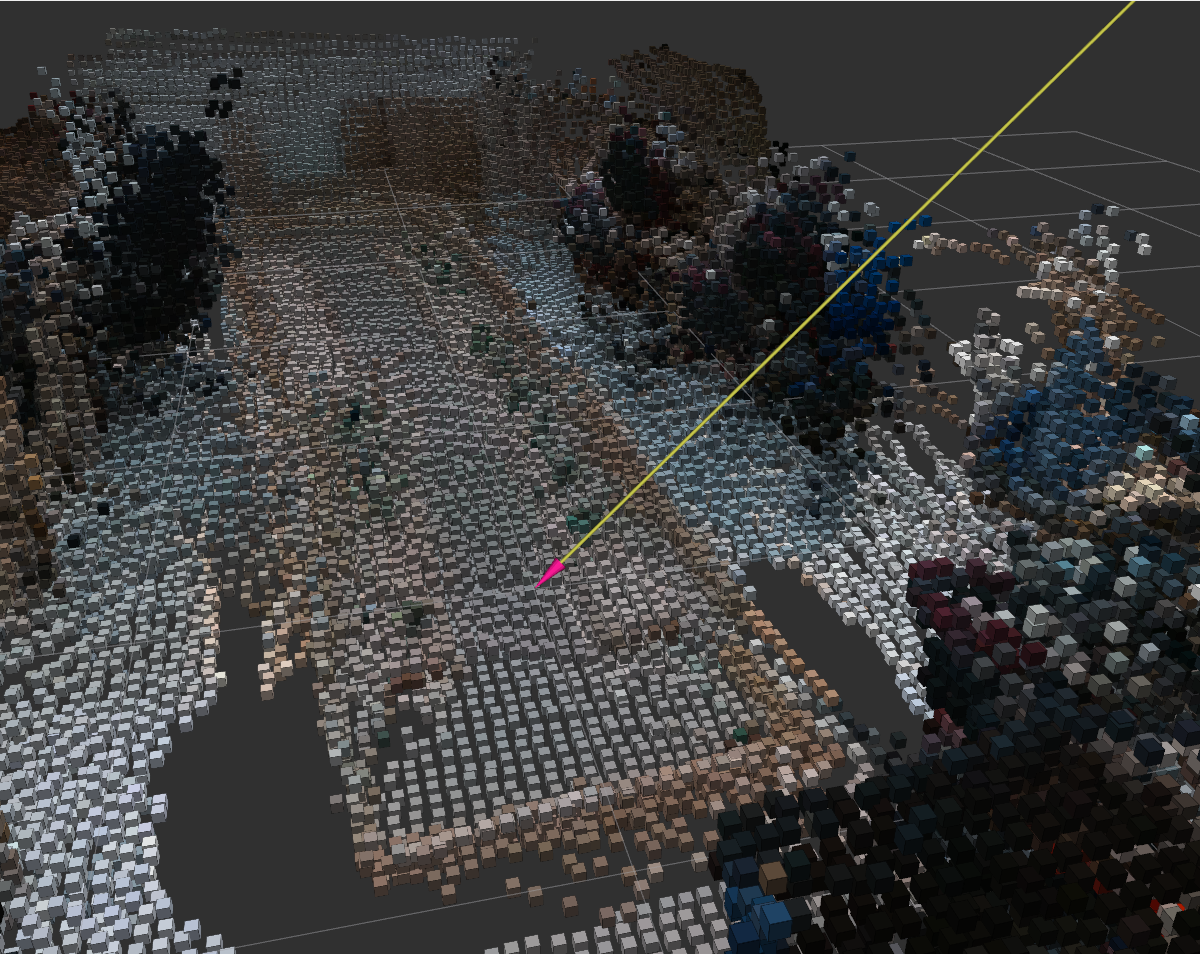}
\caption{Map Reconstruction Retrieved from RTABMAP Database}
\label{fig:slam_output}
\end{minipage}
\hfill
\begin{minipage}[b]{0.34\linewidth}
    \centering
    \includegraphics[width=\linewidth]{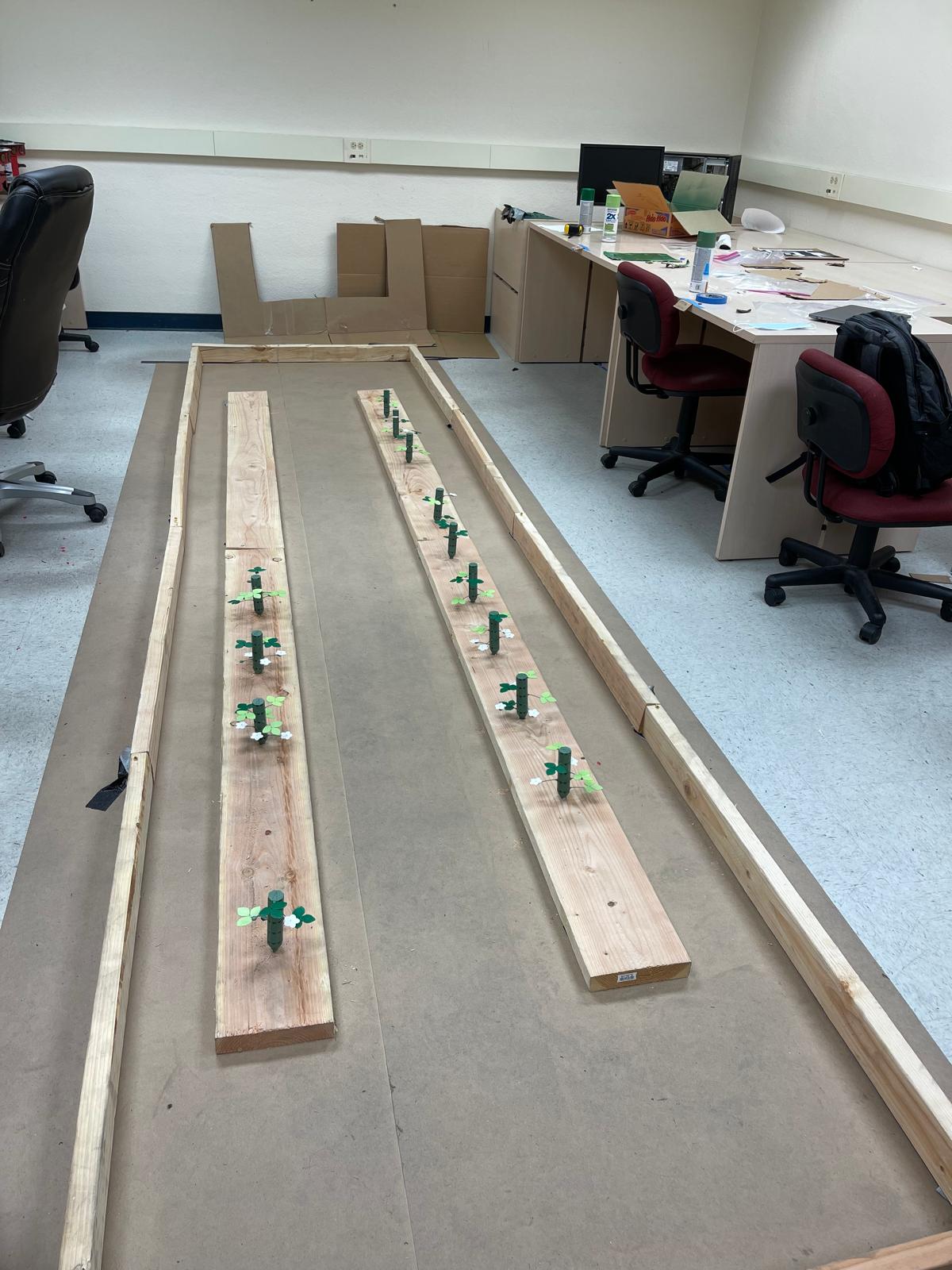}
\caption{Ground Truth of the Arena}
\label{fig:navigation_system_gt}
\end{minipage}

\end{figure*}

By breaking down the software architecture into these three main modules, we ensure a clear separation of concerns and enable efficient communication between the subsystems. The Behaviour Coordinator manages the high-level decision-making, the Navigation System handles localization and motion control, and the Plant Processing System focuses on detecting, localizing, and manipulating through plant parts.

As previously mentioned, the Armpi Pro’s Raspberry Pi is equipped with ROS Melodic, which manages manipulation-related topics along with other operational topics and services for robot hardware. The manipulation topic processes an array of integers corresponding to the number of arm joints (ranging between 0 to 1000), with each integer representing the rotation of a joint in radians.

On the ROS$2$ end, our focus is on performing inverse kinematics and obstacle avoidance. During inverse kinematics, the desired goal location ($3D$ position and orientation) is received detection and localization. We then calculate the inverse kinematics to model the poses of each joint in the arm so that the end-effector reaches the desired location. The estimated poses, expressed in radians, are scaled between $0-1000$ and passed to the ROS$1$ topic via the aforementioned REST API call.

Initially, we utilized the Robotics Toolbox (RTB) for inverse kinematics, but found that RTB sometimes estimated poses beyond the mechanical limits of the joints, leading to unrealistic and unstable maneuvers. Moreover, RTB lacked pre-defined packages for obstacle avoidance. Obstacle avoidance is crucial to prevent the arm from striking hardware components at the rear of the saral$\_$bot and to avoid collisions with the plant and its healthy structure.

Moveit2 is an easy-to-use, well-structured, stable, and reliable platform for robotic manipulation. Its built-in and ready-to-use functionalities, such as inverse kinematics and obstacle avoidance, enable realistic and stable maneuvering of the arm.

To utilize Moveit2 packages, we created the URDF (Unified Robot Description Format) file of our robot, which describes each part of our arm including links, joints, and their properties. Using the URDF file, we generate an SRDF (Semantic Robot Description Format) file along with several configuration files for collision settings, kinematics, and controllers. These files are initially used for simulation purposes. In the simulation phase, we employ ROS2 control as the controller to manipulate our arm to achieve the desired goal pose. In real-time operations, we use the Armpi Pro’s built-in packages as the controller unit.

The inverse kinematics operations are performed with ease, avoiding maneuvers that would exceed the joint limits. In addition to managing joint limitations, the arm must also avoid collisions with the computation block located at the rear and the healthy parts of the plant. To facilitate this, we added a few objects into the planning scene for visualization purposes; for instance, a cuboid object behind the arm and a cylindrical object at the center of the plant. This setup helps in planning the manipulation path to avoid these objects and successfully reach the goal pose.

%% file: 7_Appropriateness_of_Tests.tex
\section{Appropriateness of Tests and Performance Data } 

We first test each of the modules we described in this paper. Then, we test the complete competition pipeline in the simulated competition environment (See Fig~\ref{fig:arena},\ref{fig:navigation_system_gt}).

\subsection{Navigation}
In this subsection, we evaluate the performance of our navigation system. 
During our tests, our robot autonomously navigated the simulated arena by following the static set of waypoints using PID based control,  while collecting RGBD and IMU(inertial Measurement Unit) data from the Intel RealSense D435i depth camera. RTABMAP processed the incoming data in real-time, constructing a 3D map of the environment (Fig. \ref{fig:slam_output}) and estimating the robot's pose within the map.

These results demonstrate the robustness and reliability of our navigation system. With more thorough testing and validation of our navigation system in the simulated environment, we will be even more confident in its ability to perform well in the actual competition. The positive results obtained from these tests underscore the appropriateness of our chosen approach and the readiness of our robot to tackle the real challenge.

\subsection{Detection}

\begin{figure}[!htb]
\centering
\includegraphics[width=0.83\linewidth, trim={0.5cm 5cm 0 0.8cm },clip]{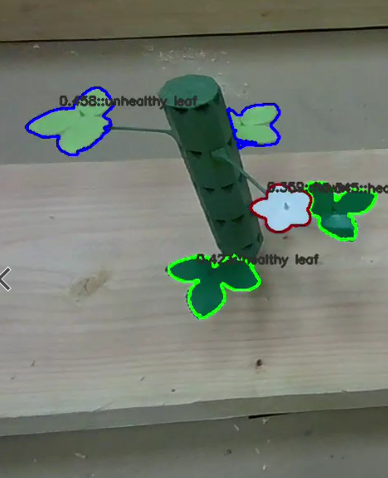}
\caption{Detected Healthy, Unhealthy Leaf-clusters, and Flowers}
\label{fig:detected_plant}
\end{figure}

In Fig~\ref{fig:detected_plant}, we show our single plant detection result.
We use accuracy as a metric for evaluating our HSV-color thresholding method for detecting healthy or unhealthy leaf clusters and flowers. 
Under different viewpoints and lighting conditions, our preliminary results show good accuracy but also exhibit false positives.
For example, non-plant objects were detected as either flower or leaf-cluster due to having similar color properties.

\begin{figure}[!htb]
\centering
\includegraphics[width=0.99\linewidth]{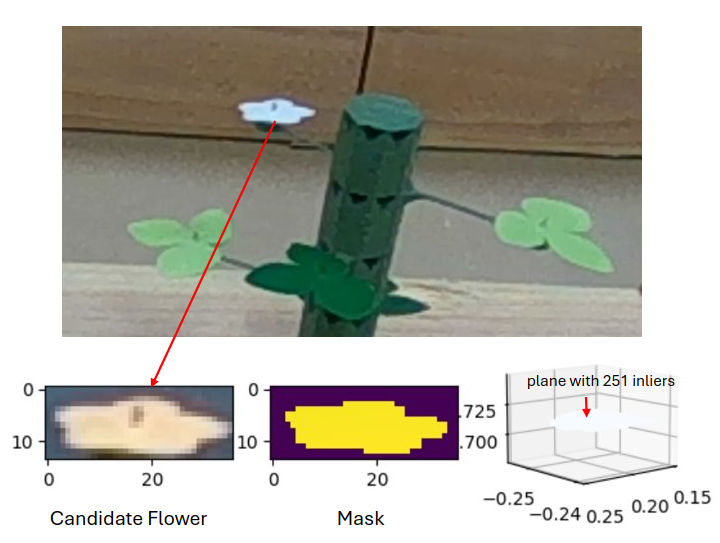}
\caption{Visualizing the Plane of a Detected Flower Object with 251 Inliers}
\label{fig:viz_planarity}
\vspace{-0.5 cm}
\end{figure}

When we include a planarity check, then it eliminates false positives. 
We visualize the plane of a detected flower object of a plant tested on our simulated test environment in Fig.~\ref{fig:viz_planarity}. 
Here, the fitted plane contains 251 inliers.

\begin{figure}[!ht]
\centering
\includegraphics[width=0.45\textwidth]{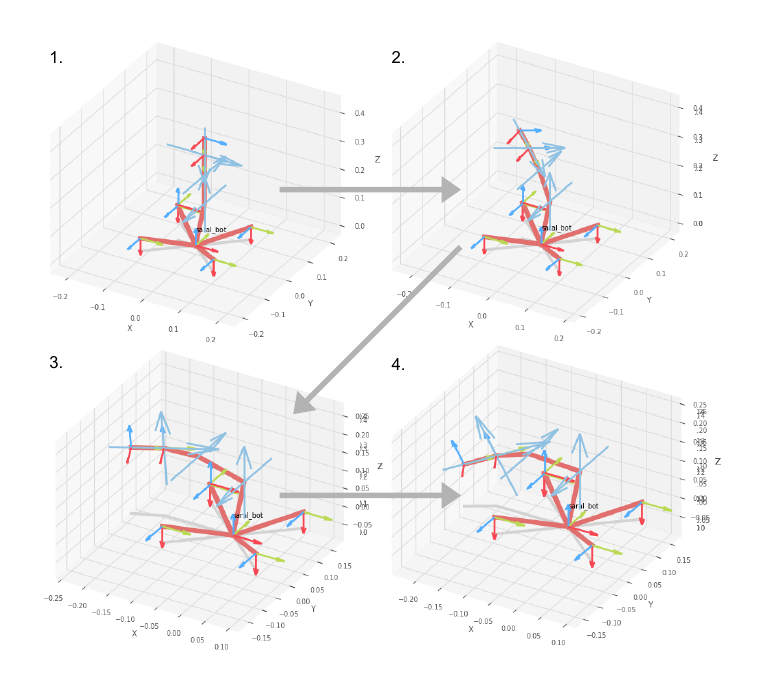}
\caption{Simulation showing the Plant Processing System Positioning Itself to Trim an Unhealthy Leaf-cluster.}
\label{fig:pprobot_rtb}
\vspace{-0.5 cm}
\end{figure}

\subsection{Manipulation}
To validate the performance of our manipulation system, we conducted a series of tests in both simulated and real-world environments. The tests aimed to assess the accuracy, reliability, and efficiency of our Inverse Kinematics (IK) solver and obstacle avoidance algorithms.
Initially, we utilized the Robotics Toolbox (RTB) Python by Dr. Peter Corke for inverse kinematics and tested the manipulation pipeline in a simulated environment. However, we encountered issues with RTB estimating poses beyond the mechanical limits of the joints, leading to unrealistic and unstable maneuvers. Moreover, RTB lacked integrated planning in presence of obstacles, which is crucial for preventing self-collisions and those with plants.
To address these limitations, we transitioned to using Moveit2, a powerful and sophisticated platform for the plant manipulation tasks. Moveit2's built-in functionalities for inverse kinematics and obstacle avoidance enable us to perform realistic and stable maneuvering of the arm.
In the simulation phase, we employed ROS2 control as the controller to manipulate the arm and achieve the desired goal pose. For all the relevant test cases that we expect to encounter throughout the course of the robot's operation, target poses are reached without violating any joint limits or self-collisions. 
To test the efficacy of our obstacle avoidance pipeline, we introduced virtual objects in the planning scene, representing the computation block at the rear of the robot and the healthy parts of the plant. We then tasked the arm with reaching target poses while avoiding these obstacles. After validating the manipulation pipeline in simulation, we deployed it on the physical robot and conducted real-world tests. Our system could successfully grasp and trim unhealthy leaves and flowers from the plant models to be used in the competition, thus demonstrating the effectiveness of our manipulation system. 
The transition from RTB to Moveit2 proved to be a significant improvement, enabling us to overcome the limitations of RTB and achieve stable and reliable maneuvers. The comprehensive testing and validation process, spanning both simulation and real-world experiments, gives us confidence in the readiness of our manipulation system for the competition.

\begin{figure}[!b]
\centering
\includegraphics[width=0.45\textwidth]{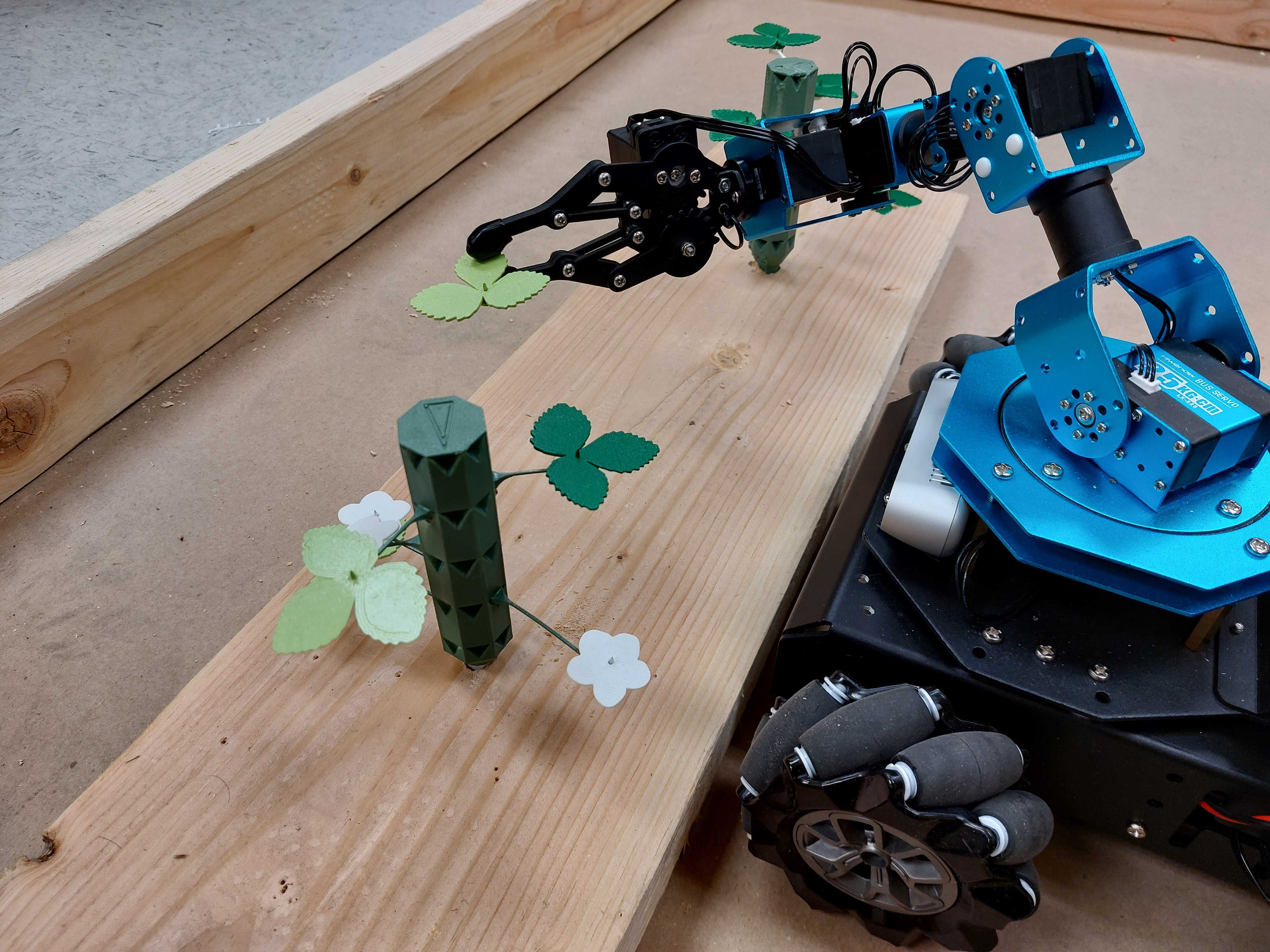}
\caption{Plant Processing System Trimming an Unhealthy Leaf-cluster with a 5-DoF Arm  and 1-DoF Lateral Movement}
\label{fig:trmming_manouver}
\end{figure}

%% file: 8_Achievement_of_Objectives.tex
\section{Achievement of Objectives 	}  

Throughout the development process of our robot, we have maintained our focus on achieving the objectives and design criteria established in Sec \ref{sec:objectives} of this report. By leveraging innovative hardware enhancements, a modular software architecture, and advanced algorithms, we have created a robot that successfully addresses the challenges posed by the competition. Now, we will discuss what we achieved from our two primary objectives.  

Our first objective was to develop a robot capable of autonomously navigating the competition arena while adhering to the specified size constraints. 
The integration of the ArmPi Pro chassis with custom hardware additions, such as the linear actuator and depth cameras, has allowed us to create a compact and maneuverable platform. The use of RTABMAP for simultaneous localization and mapping, combined with efficient path planning and motion control algorithms, enables our robot to navigate the arena with precision and reliability.

Our second objective was to achieve accurate and robust plant inspection and selective plucking/trimming. The incorporation of an RGBD camera mounted on the linear actuator at the rear of the robot provides a clear view of the plants, allowing for efficient detection and localization of leaves and flowers. Our HSV-based color thresholding and contour detection algorithms, coupled with point cloud processing, enable the robot to distinguish between healthy and unhealthy plant parts with high accuracy. The integration of Moveit2 for inverse kinematics and obstacle avoidance ensures precise and collision-free manipulation of the robot arm during the trimming process.
In addition to the competition objectives, we also aimed to create a modular and scalable system that could be easily adapted and maintained. The use of ROS2 as the primary software framework facilitates a modular and reusable codebase, allowing for the seamless integration of new features and algorithms. The separation of concerns between the Behaviour Coordinator, Navigation System, and Plant Processing System promotes a clean and maintainable software architecture. Furthermore, the utilization of Docker for environment management ensures consistent and reproducible deployments across different systems.

Throughout the development process, we've been conducting extensive tests and evaluations to validate the performance and reliability of our robot. 
The achievement of these objectives is a testament to the hard work, dedication, and technical expertise of our team. The successful integration of navigation, perception, and manipulation capabilities in a compact and efficient package demonstrates the potential for autonomous robots to revolutionize the way we approach tasks such as plant inspection and selective harvesting.
Looking ahead, we believe that the knowledge and experience gained through this project will serve as a foundation for further advancements in agricultural robotics. The modular and scalable nature of our system opens up possibilities for adapting the robot to various agricultural applications beyond the scope of the competition. By continuing to refine and expand upon the capabilities demonstrated by our robot, we aim to contribute to the development of innovative solutions that address the challenges faced by modern agriculture, ultimately promoting efficiency, sustainability, and improved crop health.